\makeatletter\def\graphicscache@inhibit{true}\makeatother
\documentclass{llncs}
\usepackage{makeidx}  %

\usepackage[utf8]{inputenc}

\usepackage[hidelinks]{hyperref}

\usepackage[numbers,sort&compress,sectionbib]{natbib}
\usepackage{url}
\usepackage{graphicx}
\usepackage[usenames,dvipsnames]{xcolor}
\usepackage[draft,inline,nomargin]{fixme}
\fxsetup{theme=color}

\usepackage{amsmath}
\usepackage{amssymb}
\usepackage{bbm}
\usepackage{textcomp}
\usepackage{siunitx}

\usepackage{booktabs}
\usepackage{threeparttable}
\usepackage{multirow}

\usepackage[capitalize]{cleveref}

\usepackage{tikz}
\usepackage{tikz-3dplot}
\usepackage[normalem]{ulem}   
\usetikzlibrary{arrows}
\usetikzlibrary{positioning,calc}
\usetikzlibrary{decorations.pathreplacing}
\usetikzlibrary{decorations.markings}
\usetikzlibrary{fit}
\usetikzlibrary{shapes.callouts}
\usetikzlibrary{shapes.geometric}
\usetikzlibrary{matrix}
\usetikzlibrary{spy}
\usepackage{setspace}

\usepackage{pgfplots}
\pgfplotsset{compat=1.9}
\usepgfplotslibrary{groupplots}
\usepgfplotslibrary{units}
\usepgfplotslibrary{colorbrewer}
\usepackage{pgfplotstable}

\usepackage{balance}

\usepackage{blindtext}

\usepackage{ifthen}
\usepackage{tabularx}
\usepackage{makecell}

\usepackage[export]{adjustbox}

\IfFileExists{graphicscache.sty}{\usepackage{graphicscache}}

\DeclareMathOperator*{\argmin}{arg\,min}

\usepackage{placeins}
\usepackage{dblfloatfix}
\newcommand{\rotateRPY}[3]%
{   \pgfmathsetmacro{\rollangle}{#1}
    \pgfmathsetmacro{\pitchangle}{#2}
    \pgfmathsetmacro{\yawangle}{#3}

    \pgfmathsetmacro{\newxx}{cos(\yawangle)*cos(\pitchangle)}
    \pgfmathsetmacro{\newxy}{sin(\yawangle)*cos(\pitchangle)}
    \pgfmathsetmacro{\newxz}{-sin(\pitchangle)}
    \path (\newxx,\newxy,\newxz);
    \pgfgetlastxy{\nxx}{\nxy};

    \pgfmathsetmacro{\newyx}{cos(\yawangle)*sin(\pitchangle)*sin(\rollangle)-sin(\yawangle)*cos(\rollangle)}
    \pgfmathsetmacro{\newyy}{sin(\yawangle)*sin(\pitchangle)*sin(\rollangle)+ cos(\yawangle)*cos(\rollangle)}
    \pgfmathsetmacro{\newyz}{cos(\pitchangle)*sin(\rollangle)}
    \path (\newyx,\newyy,\newyz);
    \pgfgetlastxy{\nyx}{\nyy};

    \pgfmathsetmacro{\newzx}{cos(\yawangle)*sin(\pitchangle)*cos(\rollangle)+ sin(\yawangle)*sin(\rollangle)}
    \pgfmathsetmacro{\newzy}{sin(\yawangle)*sin(\pitchangle)*cos(\rollangle)-cos(\yawangle)*sin(\rollangle)}
    \pgfmathsetmacro{\newzz}{cos(\pitchangle)*cos(\rollangle)}
    \path (\newzx,\newzy,\newzz);
    \pgfgetlastxy{\nzx}{\nzy};
}

\tikzset{RPY/.style={x={(\nxx,\nxy)},y={(\nyx,\nyy)},z={(\nzx,\nzy)}}}

\newcommand{\Crossblue}{$\mathbin{\tikz [x=1.ex,y=1.ex,line width=.2ex, blue] \draw (0,0) -- (1,1) (0,1) -- (1,0);}$}%
\begin{document}

\title{
YOLOPose: Transformer-based Multi-Object 6D Pose Estimation using Keypoint Regression
}

\titlerunning{YOLOPose}  %
\author{Arash Amini$^{*}$ \and Arul Selvam Periyasamy$^{*}$ \and Sven Behnke}

\institute{Autonomous Intelligent Systems, University of Bonn, Germany\\
\email{periyasa@ais.uni-bonn.de}\\
{$^{*}$Equal contribution.}
}

\maketitle

\begin{abstract}
    6D object pose estimation is a crucial prerequisite for autonomous robot manipulation applications.
    The state-of-the-art models for pose estimation are convolutional neural network (CNN)-based.
    Lately, Transformers, an architecture originally proposed for natural language processing, is achieving state-of-the-art
    results in many computer vision tasks as well. Equipped with the multi-head self-attention mechanism, Transformers
    enable simple single-stage end-to-end architectures for learning object detection and 6D object pose estimation jointly.
    In this work, we propose YOLOPose (short form for You Only Look Once Pose estimation), 
    a Transformer-based multi-object 6D pose estimation method based on keypoint regression.
    In contrast to the standard heatmaps for predicting keypoints in an image, we directly regress the keypoints.
    Additionally, we employ a learnable orientation estimation module to predict the orientation from the keypoints.
    Along with a separate translation estimation module, our model is end-to-end differentiable. Our method
    is suitable for real-time applications and achieves results comparable to state-of-the-art methods.
\end{abstract}

Autonomous robotic object manipulation in real-world scenarios depends on high-quality 6D object pose estimation.
Such object poses are also crucial in many other applications like augmented reality, autonomous navigation, and industrial bin picking.
In recent years, with the advent of convolutional neural networks (CNNs), significant progress has been made 
to boost the performance of object pose estimation methods. Due to the complex nature of the problem, 
the standard methods favor multi-stage approaches,
i.e., feature extraction followed by object detection and/or instance segmentation, target object crop extraction, and, finally,
6D object pose estimation. In contrast,
~\citet{carion2020end} introduced DETR, a Transformer-based single-stage architecture for object detection.
In our previous work~\citep{arash2021gcpr}, we extended the DETR model with the T6D-Direct architecture to perform multi-object 6D pose direct regression. Compared to multi-stage CNN-based methods that employ components like bounding box proposals, region of interest (RoI) pooling,
non-maximum suppression (NMS) to construct end-to-end differentiable pipelines, 
the T6D-Direct model learns object detection and 6D pose estimation jointly.
Taking advantage of the \textit{pleasingly parallel} nature of the Transformer architecture, the T6D-Direct model
predicts 6D poses for all the objects in an image in one forward-pass. Despite the advantages of the architecture and its
impressive performance, the overall 6D pose estimation accuracy of T6D-Direct, which directly regresses translation 
and orientation components of the 6D object poses, is inferior to state-of-the-art CNN-based methods, especially in rotation estimation.
Instead of directly regressing the translation and orientation components, the keypoint-based methods predict the 2D pixel
projection of 3D keypoints and use the P\textit{n}P algorithm to recover the 6D pose. 
We extend our T6D-Direct model to learn sparse 2D-3D correspondences.
Our proposed model performs keypoint direct regression instead of the standard heatmaps for predicting the spatial position of the keypoints
in a given RGB image and uses a multi layer perceptron (MLP) to learn the orientation component of 6D object pose from the keypoints. Another independent MLP serves as the translation estimator.
\noindent
In short, our contributions include:
\begin{enumerate}
 \item A Transformer-based real-time single-stage model for multi-object monocular 6D pose estimation using keypoint regression,
 \item a learnable rotation estimation module to estimate object orientation from a set of keypoints to develop an end-to-end differentiable architecture for pose estimation, and
 \item results comparable to state-of-the-art pose estimation methods on the YCB-Video dataset while being capable of real-time inference.
\end{enumerate}

\begin{figure}[t]
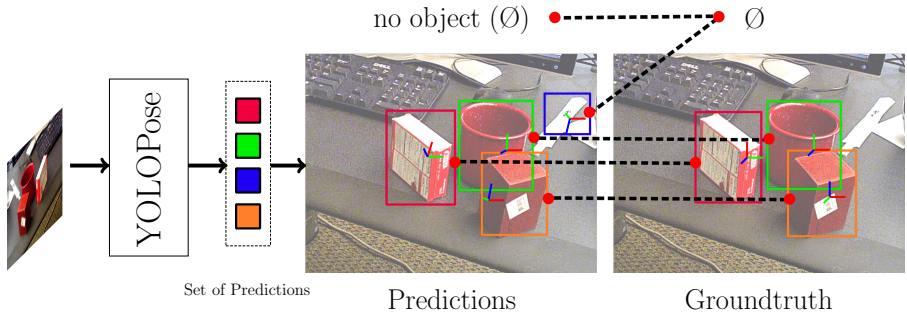

        \centering
          \resizebox{.99\linewidth}{!}{%
\begin{tikzpicture}

\node (canvas) at (7, 0) {\includegraphics[width=17cm]{figures/pipeline/pipeline.png}};

\node[canvas is zy plane at x=0] (input) at (-6.6, -1.2, 0) {\includegraphics[width=0.3\textwidth]{figures/architecture/input.png}};

\node [align=center] at (4, 3.1){\huge no object (\O)};
\node [align=center] at (11.7, 3.15){\huge \O};

\draw[->,  thick, line width=1mm] (-2.7,-0.6) -- (-1.7,-0.6);
\draw[->,  thick, line width=1mm] (-5.7,-0.6) -- (-4.7,-0.6);
\draw[->,  thick, line width=1mm] (-0.6,-0.6) -- (0.3,-0.6);

\draw[draw=black, rotate around={90:(-3.5,-0.2)}] (-6.2,-1) rectangle ++(4.5, 2);
\node [align=center, rotate=90] at (-3.7, -0.7){\Huge YOLOPose};

\node [align=center] at (-1.2, -3.75){\large Set of Predictions};

\node [align=center] at (4, -4){\huge Predictions};
\node [align=center] at (11.8, -4){\huge Groundtruth};

\node(dummy1) at (-0.85, -2.4) {};
\node(dummy2) at (-1.5,  1.3) {};
\node[draw,   dash pattern={on 2pt off 1pt on 2pt off 1pt}, fit=(dummy1)(dummy2)] {};

\rotateRPY{-30}{00}{0}
    \begin{scope}[RPY]
      \coordinate (O1) at (5.2, -1.4, 0);
\draw[red, thick,-, line width=0.5mm] ( $ (O1) + (0,0,0) $) -- ( $ (O1) + (0.4,0,0) $) node[anchor=north east]{};
\draw[blue, thick,-, line width=0.5mm] ( $ (O1) + (0,0,0) $)  -- ( $ (O1) +(0,0.4,0) $) node[anchor=north west]{};
\draw[green, thick,-, line width=0.5mm] ( $ (O1) + (0,0,0) $)  -- ( $ (O1) +(0,0,0.4) $) node[anchor=south]{}; 
    \end{scope}

\rotateRPY{-10}{00}{0}
    \begin{scope}[RPY]
       \coordinate (O1) at (5.4, -0.2, 0);
\draw[red, thick,-, line width=0.5mm] ( $ (O1) + (0,0,0) $) -- ( $ (O1) + (0.4,0,0) $) node[anchor=north east]{};
\draw[green, thick,-, line width=0.5mm] ( $ (O1) + (0,0,0) $)  -- ( $ (O1) +(0,0.4,0) $) node[anchor=north west]{};
\draw[blue, thick,-, line width=0.5mm] ( $ (O1) + (0,0,0) $)  -- ( $ (O1) +(0,0,0.4) $) node[anchor=south]{}; 
    \end{scope}

\rotateRPY{50}{00}{0}
    \begin{scope}[RPY]
       \coordinate (O1) at (7.5, 1.7, 0);
\draw[red, thick,-, line width=0.5mm] ( $ (O1) + (0,0,0) $) -- ( $ (O1) + (0.4,0,0) $) node[anchor=north east]{};
\draw[green, thick,-, line width=0.5mm] ( $ (O1) + (0,0,0) $)  -- ( $ (O1) +(0,0.4,0) $) node[anchor=north west]{};
\draw[blue, thick,-, line width=0.5mm] ( $ (O1) + (0,0,0) $)  -- ( $ (O1) +(0,0,0.4) $) node[anchor=south]{}; 
    \end{scope}

\rotateRPY{-50}{00}{0}
    \begin{scope}[RPY]
      \coordinate (O1) at (3.5, -0.4, 0);
\draw[green, thick,-, line width=0.5mm] ( $ (O1) + (0,0,0) $) -- ( $ (O1) + (0.4,0,0) $) node[anchor=north east]{};
\draw[red, thick,-, line width=0.5mm] ( $ (O1) + (0,0,0) $)  -- ( $ (O1) +(0,0.4,0) $) node[anchor=north west]{};
\draw[blue, thick,-, line width=0.5mm] ( $ (O1) + (0,0,0) $)  -- ( $ (O1) +(0,0,0.4) $) node[anchor=south]{}; 
    \end{scope}

\rotateRPY{0}{0}{0}
    \begin{scope}[RPY]
       \coordinate (O1) at (11.5, -0.4, 0);
\draw[green, thick,-, line width=0.5mm] ( $ (O1) + (0,0,0) $) -- ( $ (O1) + (0.4,0,0) $) node[anchor=north east]{};
\draw[red, thick,-, line width=0.5mm] ( $ (O1) + (0,0,0) $)  -- ( $ (O1) +(0,0.4,0) $) node[anchor=north west]{};
\draw[blue, thick,-, line width=0.5mm] ( $ (O1) + (0,0,0) $)  -- ( $ (O1) +(0,0,0.4) $) node[anchor=south]{}; 
    \end{scope}

\rotateRPY{0}{0}{0}
    \begin{scope}[RPY]
       \coordinate (O1) at (13.2, -0.2, 0);
\draw[red, thick,-, line width=0.5mm] ( $ (O1) + (0,0,0) $) -- ( $ (O1) + (0.4,0,0) $) node[anchor=north east]{};
\draw[green, thick,-, line width=0.5mm] ( $ (O1) + (0,0,0) $)  -- ( $ (O1) +(0,0.4,0) $) node[anchor=north west]{};
\draw[blue, thick,-, line width=0.5mm] ( $ (O1) + (0,0,0) $)  -- ( $ (O1) +(0,0,0.4) $) node[anchor=south]{}; 
    \end{scope}
    
\rotateRPY{0}{0}{0}
    \begin{scope}[RPY]
       \coordinate (O1) at (13.6, -1.4, 0);
\draw[red, thick,-, line width=0.5mm] ( $ (O1) + (0,0,0) $) -- ( $ (O1) + (0.4,0,0) $) node[anchor=north east]{};
\draw[blue, thick,-, line width=0.5mm] ( $ (O1) + (0,0,0) $)  -- ( $ (O1) +(0,0.4,0) $) node[anchor=north west]{};
\draw[green, thick,-, line width=0.5mm] ( $ (O1) + (0,0,0) $)  -- ( $ (O1) +(0,0,0.4) $) node[anchor=south]{}; 
    \end{scope}

\end{tikzpicture} 
}
          \caption{Proposed YOLOPose approach. Our model predicts a set with a fixed cardinality $N$. Each element in the set corresponds to an object prediction and after predicting all the objects in
          the given input image, the rest of the elements are padded with \text{\O} as no object predictions. The predicted and the groundtruth sets are matched using bipartite matching and the model is trained to minimize the Hungarian loss
          between the matched pairs. Our model is end-to-end differentiable.
          }
          \label{fig:pipeline}
\end{figure}

\section{Related Work}
\subsection{RGB Object Pose Estimation}
\label{sec:models}

The recent significant progress in the task of 6D object pose estimation from RGB images is driven---like in many computer vision tasks---by deep learning methods. 
The methods for the object pose estimation from RGB images can be broadly classified into three major categories, namely direct regression methods, keypoint-based methods, and refinement-based methods.
Direct regression methods formulate the problem of pose estimation as a regression of continuous translation and rotation components, 
whereas keypoint-based methods predict the location of projection of some of the specific keypoints or the 3D coordinates of the visible pixels of an object 
in an image and use the P\textit{n}P algorithm to recover the 6D  pose from the estimated 2D-3D correspondences. 
The P\textit{n}P algorithm is used in conjunction with RANSAC to improve the robustness of the pose estimation.  

Some examples for direct regression methods include~\citep{xiang2017posecnn,Wang_2021_GDRN,periyasamy2018pose,arash2021gcpr} and 
examples for keypoint-based methods include~\citep{rad2017bb8,tekin2018real,hu2019segmentation,peng2019pvnet,hu2020single}.
One important detail to note is that, except for~\citep{arash2021gcpr} all the other methods use multi-staged CNNs. 
In the first stage, the model performs object detection and/or 
semantic or instance segmentation to detect the objects in the given RGB image. Using the object detections from the first stage, a 
crop containing the target object is extracted. In the second stage, the model predicts the 6D pose of the target object from the image crop. 
In terms of the 6D pose prediction accuracy, keypoint-based methods perform considerably better than the direct regression methods~\citep{hodavn2020bop}, 
though this performance gap is shrinking~\citep{arash2021gcpr}. 

The third category of the pose estimation methods is refinement-based. These methods formulate the task of pose estimation as iterative pose refinement, 
i.e., the target object is rendered according to the current pose estimate, and a model is trained to estimate a pose update that minimizes the pose error between the groundtruth and the current pose prediction. 
Refinement-based methods~\citep{labbe2020, manhardt2018deep, Shao_2020_CVPR, periyasamy2019refining, li2018deepim} achieve the highest pose prediction accuracy among all categories~\citep{hodavn2020bop}.

\subsection{Learned P\textit{n}P}
Given a set of 3D keypoints and their corresponding 2D projections and the camera intrinsics, the P\textit{n}P algorithm is used to recover the 6D object pose.
The standard P\textit{n}P algorithm~\citep{Gao2003CompleteSC} and its variant EP\textit{n}P~\citep{lepetit2009epnp} are used in combination
with RANSAC to improve the robustness against outliers. Both P\textit{n}P and RANSAC are not trivially differentiable. In order to
realize an end-to-end differentiable pipeline for the 6D object pose estimation,~\citet{Wang_2021_GDRN}, and~\citet{hu2020single} proposed a learning-based P\textit{n}P  module. 
Similarly,~\citet{Li_2021_CVPR} introduced a learnable
3D Lifter module to estimate vehicle orientation. Lately,~\citet{chen2020end} proposed to differentiate  P\textit{n}P using the implicit function theorem. 
Although a generic differentiable P\textit{n}P 
has many potentials, due to the overhead incurred during training, we opt for a simple MLP that estimates the orientation component given the 2D keypoints.
\begin{figure*}
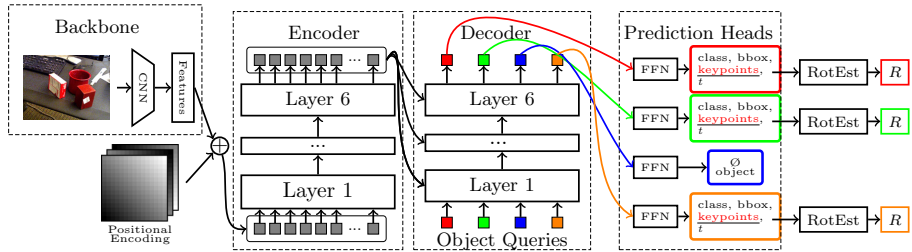

        \centering
          \resizebox{.99\linewidth}{!}{%
\begin{tikzpicture}
\pgfmathsetmacro{\EITH}{1.15}
\node (inimage) at (0, 0.7) {\includegraphics[width=1.4cm]{figures/architecture/input.png}};
\node(cnn) at (1.25, 0.7) [trapezium, trapezium angle=55, rotate=-90, minimum width=10mm, draw, thick, ] {\tiny CNN};
\node (feat) at (1.9, 0.7) [draw,thick,minimum width=0.2cm,minimum height=0.2cm, rotate=-90] {\tiny Features};
\coordinate (concat) at (2.5, -0.25);
\node[circle,draw=black, thick, fill=white, inner sep=0pt,minimum size=8pt] (concatcircle) at (concat) {};
\node (backbone) at (0.5, 1.7) [] {\footnotesize Backbone};
\node (posenc) at (1.2, -1.7) [align=center, font=\tiny\linespread{0.7}\selectfont]{\tiny Positional \\ \tiny Encoding};
\draw (concat)[very thick] node[rotate=0]{+};
\node (poscode) at (1.2, -0.85) {\includegraphics[width=1.25cm]{figures/architecture/positional_encoding.png}};

\draw [thick, ->](inimage.east) -- (cnn.south);
\draw [thick, ->](cnn.north) -- (feat.south);
\draw [thick, ->](feat.north) -- (concatcircle.135);
\draw [thick, ->](poscode.east) -- (concatcircle.225);

\node(dummy0) at (0, 1.7) {};
\node[draw,  dash pattern={on 2pt off 1pt on 2pt off 1pt},fit=(inimage) (cnn) (feat) (backbone) (dummy0)] {};

\node (ib1) at (3.15, -1.6) [draw, fill=gray,minimum  size=2 pt, scale=0.75] {};
\node (ib2) at (3.4, -1.6) [draw, fill=gray,minimum  size=2 pt, scale=0.75] {};
\node (ib3) at (3.65, -1.6) [draw, fill=gray,minimum  size=2 pt, scale=0.75] {};
\node (ib4) at (3.9, -1.6) [draw, fill=gray,minimum  size=2 pt, scale=0.75] {};
\node (ib5) at (4.15, -1.6) [draw, fill=gray,minimum  size=2 pt, scale=0.75] {};
\node (ib6) at (4.4, -1.6) [draw, fill=gray,minimum  size=2 pt, scale=0.75] {};
\node (ib7) at (4.7, -1.6) [minimum  size=2 pt, scale=0.75] {...};
\node (ib8) at (5., -1.6) [draw, fill=gray,minimum  size=2 pt, scale=0.75]{};

\node (tenc1) at (4.1, -1) [draw,thick,minimum width=2.5cm, align=center] {Layer 1};
\node (tenc2) at (4.1, -0.25) [draw,thick,minimum width=2.5cm,align=center] {\small ...};
\node (tenc) at (4.1, 0.5) [draw,thick,minimum width=2.5cm, align=center] {\small Layer 6};

\draw[thick, ->] (ib1.north) -- (ib1.north|-tenc1.south);
\draw[thick, ->] (ib2.north) -- (ib2.north|-tenc1.south);
\draw[thick, ->] (ib3.north) -- (ib3.north|-tenc1.south);
\draw[thick, ->] (ib4.north) -- (ib4.north|-tenc1.south);
\draw[thick, ->] (ib5.north) -- (ib5.north|-tenc1.south);
\draw[thick, ->] (ib6.north) -- (ib6.north|-tenc1.south);
\draw[thick, ->] (ib8.north) -- (ib8.north|-tenc1.south);

\draw[thick, ->] (tenc1.north) -- (tenc1.north|-tenc2.south);
\draw[thick, ->] (tenc2.north) -- (tenc2.north|-tenc.south);

\node(encfit)[draw, rounded corners=.05cm, fit=(ib1) (ib8) ] {};

\draw [thick, ->]    (concatcircle) to[out=-75, in=190] (encfit.west);

\node (it1) at (3.15, \EITH) [draw, fill=gray,minimum  size=2 pt, scale=0.75] {};
\node (it2) at (3.4,  \EITH) [draw, fill=gray,minimum  size=2 pt, scale=0.75] {};
\node (it3) at (3.65, \EITH) [draw, fill=gray,minimum  size=2 pt, scale=0.75] {};
\node (it4) at (3.9, \EITH) [draw, fill=gray,minimum  size=2 pt, scale=0.75] {};
\node (it5) at (4.15, \EITH) [draw, fill=gray,minimum  size=2 pt, scale=0.75] {};
\node (it6) at (4.4, \EITH) [draw, fill=gray,minimum  size=2 pt, scale=0.75] {};
\node (it7) at (4.7, \EITH) [minimum  size=2 pt, scale=0.75] {...};
\node (it8) at (5., \EITH) [draw, fill=gray,minimum  size=2 pt, scale=0.75]{};
\node(encfittop)[draw, rounded corners=.05cm, fit=(it1) (it8) ] {};

\draw[thick, <-] (it1.south) -- (it1.north|-tenc.north);
\draw[thick, <-] (it2.south) -- (it2.north|-tenc.north);
\draw[thick, <-] (it3.south) -- (it3.north|-tenc.north);
\draw[thick, <-] (it4.south) -- (it4.north|-tenc.north);
\draw[thick, <-] (it5.south) -- (it5.north|-tenc.north);
\draw[thick, <-] (it6.south) -- (it6.north|-tenc.north);
\draw[thick, <-] (it8.south) -- (it8.north|-tenc.north);

\node (encoder) at (4.2, 1.6) [] {\small Encoder};
\node(dummy1) at (5, -1.7) {};
\node(dummy2) at (5,  1.7) {};
\node[draw,   dash pattern={on 2pt off 1pt on 2pt off 1pt}, fit=(it1) (ib8) (tenc)(dummy2)(dummy1)] {};

\node (ob1) at (6.2, -1.5) [draw, fill=red,minimum  size=2 pt, scale=0.75] {};
\node (ob2) at (6.8, -1.5) [draw, fill=green,minimum  size=2 pt, scale=0.75] {};
\node (ob3) at (7.4, -1.5) [draw, fill=blue,minimum  size=2 pt, scale=0.75] {};
\node (ob4) at (8., -1.5) [draw, fill=orange,minimum  size=2 pt, scale=0.75]{};
\node (ot1) at (6.2, \EITH) [draw, fill=red,minimum  size=2 pt, scale=0.75] {};
\node (ot2) at (6.8, \EITH) [draw, fill=green,minimum  size=2 pt, scale=0.75] {};
\node (ot3) at (7.4, \EITH) [draw, fill=blue,minimum  size=2 pt, scale=0.75] {};
\node (ot4) at (8.,   \EITH) [draw, fill=orange,minimum  size=2 pt, scale=0.75]{};

\node (tdec) at (7.1, -0.9) [draw,thick,minimum width=2.5cm, align=center] { \small Layer 1};
\node (tdec1) at (7.1, -0.2) [draw,thick,minimum width=2.5cm, align=center] { ... };
\node (tdec2) at (7.1, .5) [draw,thick,minimum width=2.5cm, align=center] { \small Layer 6};

\draw[thick, ->] (tdec1.north) -- (tdec1.north|-tdec2.south);
\draw[thick, ->] (tdec.north) -- (tdec.north|-tdec1.south);

\draw[thick, <-] (ot1.south) -- (ot1.north|-tdec2.north);
\draw[thick, <-] (ot2.south) -- (ot2.north|-tdec2.north);
\draw[thick, <-] (ot3.south) -- (ot3.north|-tdec2.north);
\draw[thick, <-] (ot4.south) -- (ot4.north|-tdec2.north);

\draw[thick, ->] (ob1.north) -- (ob1.north|-tdec.south);
\draw[thick, ->] (ob2.north) -- (ob2.north|-tdec.south);
\draw[thick, ->] (ob3.north) -- (ob3.north|-tdec.south);
\draw[thick, ->] (ob4.north) -- (ob4.north|-tdec.south);

\draw [thick, ->]    (encfittop.east) to[out=60, in=150] (tdec.west);

\draw [thick, ->]    (encfittop.east) to[out=60, in=150] (tdec.west);

\draw [thick, ->]    (encfittop.east) to[out=60, in=150] (tdec1.west);

\draw [thick, ->]    (encfittop.east) to[out=60, in=150] (tdec2.west);

\node (encoder) at (7, 1.6) [] {\small Decoder};
\node (encoder) at (7.1, -1.8) [] {\small Object Queries};
\node(dummy3) at (5.9, -1.7) {};
\node(dummy4) at (8.3,  1.7) {};
\node[draw,   dash pattern={on 2pt off 1pt on 2pt off 1pt}, fit=(dummy3)(dummy4)] {};

\node (feat1) at (9.6, 1) [draw,thick,minimum width=0.2cm,minimum height=0.2cm] {\tiny FFN};
\node (feat2) at (9.6, 0.2) [draw,thick,minimum width=0.2cm,minimum height=0.2cm] {\tiny FFN};
\node (feat3) at (9.6, -1.4) [draw,thick,minimum width=0.2cm,minimum height=0.2cm] {\tiny FFN};
\node (feat4) at (9.6, -0.6) [draw,thick,minimum width=0.2cm,minimum height=0.2cm] {\tiny FFN};

\node (feat5) at (10.9, 1) [draw=red,very thick,minimum width=0.15cm,minimum height=0.2cm, align=left, font=\tiny\linespread{0.7}\selectfont,rounded corners=.05cm,] {\tiny class, bbox, \\ \uline{\tiny \textcolor{red}{keypoints}},\\ $\textit{t}$};
\node (feat6) at (10.9, 0.2) [draw=green,very thick,minimum width=0.2cm,minimum height=0.2cm, align=left,font=\tiny\linespread{0.7}\selectfont,rounded corners=.05cm,] {\tiny class, bbox, \\ \uline{\tiny \textcolor{red}{keypoints}},\\ $\textit{t}$};
\node (feat7) at (10.9, -1.4) [draw=orange,very thick,minimum width=0.2cm,minimum height=0.2cm, align=left, font=\tiny\linespread{0.7}\selectfont, rounded corners=.05cm,] {\tiny class, bbox, \\ \uline{\tiny \textcolor{red}{keypoints}},\\ $\textit{t}$};
\node (feat8) at (10.9, -0.6) [draw=blue,very thick,minimum width=0.2cm,minimum height=0.2cm, align=center, font=\tiny\linespread{0.7}\selectfont, rounded corners=.05cm,] {\tiny \O \\ \tiny object};
\node(dummy5) at (9.25, -1.7) {};
\node(dummy6) at (11.4,  1.7) {};
\node (encoder) at (10.3, 1.6) [] {\small Prediction Heads};
\node[draw,   dash pattern={on 2pt off 1pt on 2pt off 1pt}, fit=(dummy5)(dummy6)] {};

\draw [thick, ->, red]    (ot1.north) to[out=90, in=150] (feat1.west);
\draw [thick, ->, green]    (ot2.north) to[out=45, in=140] (feat2.west);
\draw [thick, ->, blue]    (ot3.north) to[out=40, in=130] (feat4.west);
\draw [thick, ->, orange]    (ot4.north) to[out=30, in=150] (feat3.west);

\node (pnp1) at (12.5, 0.95) [draw,thick,minimum width=0.2cm,minimum height=0.2cm] {\scriptsize  RotEst};
\node (pnp2) at (12.5, 0.15) [draw,thick,minimum width=0.2cm,minimum height=0.2cm] {\scriptsize RotEst};
\node (pnp3) at (12.5, -1.45) [draw,thick,minimum width=0.2cm,minimum height=0.2cm] {\scriptsize RotEst};

\node (pose1) at (13.5, 0.95) [draw=red,thick,minimum width=0.2cm,minimum height=0.2cm] {\scriptsize $R$};
\node (pose2) at (13.5, 0.15) [draw=green,thick,minimum width=0.2cm,minimum height=0.2cm] {\scriptsize $R$};
\node (pose3) at (13.5, -1.45) [draw=orange,thick,minimum width=0.2cm,minimum height=0.2cm] {\scriptsize $R$};

\draw [thick, ->, ]      (11.5, 0.95) -- (pnp1) ;
\draw [thick, ->, ]      (11.5, 0.15) -- (pnp2) ;
\draw [thick, ->, ]      (11.5, -1.45) -- (pnp3) ;

\draw [thick, ->, ]     (pnp1.east) -- (pose1);
\draw [thick, ->, ]     (pnp2.east) -- (pose2);
\draw [thick, ->, ]     (pnp3.east) -- (pose3);

\draw [thick, ->, ]     (feat1.east) -- (feat5);
\draw [thick, ->, ]     (feat2.east) -- (feat6);
\draw [thick, ->, ]     (feat4.east) -- (feat8);
\draw [thick, ->, ]     (feat3.east) -- (feat7);

\end{tikzpicture}
}
          \caption{YOLOPose architecture in detail. Given an RGB input image, we extract features using the standard ResNet model. 
          The extracted features are supplemented with positional encoding and provided as input to the Transformer encoder.
          The encoder module consists of 6 standard encoder layers with skip connections.
          The output of the encoder module is provided to the decoder module along with $N$ object queries and the decoder module
          also consists of 6 standard decoder layers with skip connections generating $N$ output embeddings. 
          The output embeddings are processed with FFNs to generate a set of $N$ elements in parallel. Each element in the set 
          is a tuple consisting of bounding box, class probability, translation and interpolated bounding box keypoints.
          A learnable rotation estimation module is employed to estimate object orientation $R$ from the predicted keypoints.
          }
          \label{fig:architecture}
\end{figure*}

\begin{figure}
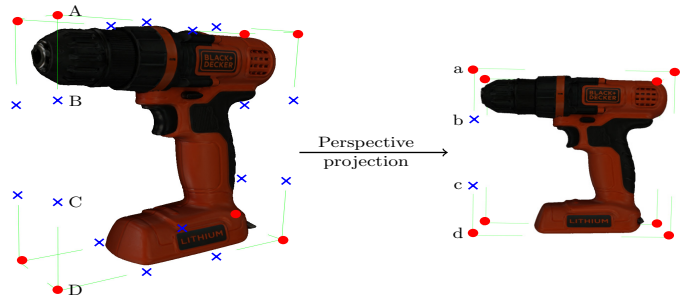

    \centering
      \resizebox*{.75\linewidth}{4cm}{%
\begin{tikzpicture}

\node (canvas) at (0, 0) {\includegraphics[width=5cm]{figures/IBB/obj.png}};
\node (canvas) at (7, 0) {\includegraphics[width=3.5cm]{figures/IBB/projected.png}};

\node[circle,fill=red,inner sep=0pt,minimum size=5pt] (a) at (-2.3,-2.275) {};
\node[circle,fill=red,inner sep=0pt,minimum size=5pt] (a) at (-2.375, 2.775) {};

\node[circle,fill=red,inner sep=0pt,minimum size=5pt] (a) at (-1.7,-2.9) {};
\node[circle,fill=red,inner sep=0pt,minimum size=5pt] (a) at (-1.7, 2.9) {};

\node[circle,fill=red,inner sep=0pt,minimum size=5pt] (a) at (2.1, -1.85) {};
\node[circle,fill=red,inner sep=0pt,minimum size=5pt] (a) at (2.35, 2.5) {};

\node[circle,fill=red,inner sep=0pt,minimum size=5pt] (a) at (1.3, -1.3) {};
\node[circle,fill=red,inner sep=0pt,minimum size=5pt] (a) at (1.45, 2.5) {};

\node[circle,fill=red,inner sep=0pt,minimum size=5pt] (a) at (5.5, -1.45) {};
\node[circle,fill=red,inner sep=0pt,minimum size=5pt] (a) at (5.5, 1.55) {};

\node[circle,fill=red,inner sep=0pt,minimum size=5pt] (a) at (5.3, -1.7) {};
\node[circle,fill=red,inner sep=0pt,minimum size=5pt] (a) at (5.3, 1.75) {};

\node[circle,fill=red,inner sep=0pt,minimum size=5pt] (a) at (8.6, -1.75) {};
\node[circle,fill=red,inner sep=0pt,minimum size=5pt] (a) at (8.7, 1.7) {};

\node[circle,fill=red,inner sep=0pt,minimum size=5pt] (a) at (8.4, -1.5) {};
\node[circle,fill=red,inner sep=0pt,minimum size=5pt] (a) at (8.4, 1.5) {};

\node at (-2.4, 1.)  {\Crossblue};
\node at (-2.375, -0.9)  {\Crossblue};

\node at (-1.7, 1.1)  {\Crossblue};
\node at (-1.7, -1.05)  {\Crossblue};

\node at (1.45, 1.)  {\Crossblue};
\node at (1.4, -0.55)  {\Crossblue};

\node at (2.275, 1.1)  {\Crossblue};
\node at (2.15,  -0.6)  {\Crossblue};

\node at(0.975, 2.65)  {\Crossblue};
\node at (-0.2,  2.75)  {\Crossblue};

\node at(0.575, 2.575)  {\Crossblue};
\node at (-0.8,  2.675)  {\Crossblue};

\node at(0.975, -2.2)  {\Crossblue};
\node at (-0.2,  -2.525)  {\Crossblue};
\node at(0.4, -1.6)  {\Crossblue};
\node at (-1.,  -1.9)  {\Crossblue};

\node at (5.32, 0.7) [red] {\Crossblue};
\node at (5.3, -0.7)  {\Crossblue};

\node(dummy1) at (2.25, 0) {};
\node(dummy2) at (5, 0) {};
\draw [thick, ->, ]      (dummy1) -- (dummy2) ;

\node  at (3.5, 0.2) [,minimum width=2.5cm, align=center] { \small Perspective};
\node  at (3.5, -0.2) [,minimum width=2.5cm, align=center] { \small projection};

\node  at (-1.4 , 3)[thick] {\small A};
\node  at (-1.4 , 1.1)[thick] {\small B};
\node  at (-1.4 , -1.05)[thick] {\small C};
\node  at (-1.4 , -2.9)[thick] {\small D};

\node  at (5.05 , 1.75)[thick] {\small a};
\node  at (5.05 , 0.7)[thick] {\small b};
\node  at (5.05 , -0.7)[thick] {\small c};
\node  at (5.05 , -1.7)[thick] {\small d};
\end{tikzpicture} 
}
      \caption{Interpolated bounding box points. Bounding box points are indicated with red dots, and the interpolated points are indicated with blue crosses. The cross-ratio of every four collinear points is preserved during perspective projection,e.g., the cross-ratio the points A, B, C, and D remains the same in 3D and, after perspective projection, in 2D.
    }
      \label{fig:IBB}
\end{figure}

\section{Method}

\subsection{Multi-Object Keypoint Regression as Set Prediction}
\label{sec:set}
Following DETR~\citep{carion2020end} and T6D-Direct~\citep{arash2021gcpr}, we formulate the problem of keypoint regression as a set prediction problem. 
Given an RGB input image, our model outputs a set of elements with a fixed cardinality $N$. Each element in the set is a tuple containing 2D bounding boxes, class probability, 
translation, and keypoints. The 2D bounding boxes are represented with the center coordinates, height, and width proportional to
the image size. 
The class probability is predicted using a softmax function. 
We regress the translation component directly, where our translation representation follows~\citet{xiang2017posecnn}.
The exact choice of the keypoints is discussed in~\cref{sec:keypoints}.
The number of objects present in an image varies. Therefore, to enable output sets with fixed cardinality, 
we choose $N$ to be larger than the expected maximum number of objects in an image in the dataset and introduce a null object class \text{\O}. The \text{\O} class is analogous to the background class used in
semantic segmentation models. In addition to predicting the corresponding classes for objects present in the image, our model is trained to predict \text{\O}
class for the rest of the elements in the set.

A set is an unordered collection of elements. 
To facilitate comparing the groundtruth set and the predicted set, we use bipartite matching~\citep{kuhn1955hungarian, stewart2016end, carion2020end} 
to find the permutation of the predicted elements
that minimizes the matching cost. Given the \text{\O} class padded groundtruth set $y$ of cardinality $N$ containing labels ${y}_1, {y}_2, ..., {y}_n$ for $n$ elements,
the predicted set denoted by $\hat{y}$, we search for the optimal permutation  %
$\hat{\sigma}$ among the possible permutations $\sigma \in \mathfrak{S}_{N}$ 
that minimizes the matching cost $\mathcal{L}_{match}$.
Formally,
\begin{equation}
\hat{\sigma} = \argmin_{\sigma \in \mathfrak{S}_{N}} \sum_i^N  \mathcal{L}_{match}(y_i, \hat{y}_{\sigma(i)}).
\end{equation}

Although each element of the set is a tuple containing four components, bounding box, class probability, translation, and keypoints, we use only the bounding box and class probability
components to define the cost function. In practice, omitting the other components in the cost function definition does not hinder the model's 
ability in learning to predict the keypoints.

\subsection{Keypoints Representation}
\label{sec:keypoints}
An obvious choice for 3D keypoints is the eight corners of the 3D bounding box~\citep{oberweger2018}.
~\citet{peng2019pvnet} argued that predicting the projection of 3D bounding boxes of an object might be difficult for a CNN-based model since the projection might lie outside of
the object boundary in the RGB image. To alleviate this issue, they proposed to use the Farthest Point Sampling algorithm (FPS) 
to sample eight keypoints on the surface of the object meshes.
~\citet{Li_2021_CVPR} defined the 3D representation of an object as sparse interpolated bounding boxes (IBBs), depicted in~\cref{fig:IBB}, and exploited the property of perspective projection that the cross-ratio of every four collinear points in 3D (A, B, C, and D as illustrated in~\cref{fig:IBB}) is preserved under perspective 
projection in 2D~\citep{hartley_zisserman_2004}.
The cross-ratio consistency is enforced by an additional component in the loss function that the model learns to minimize during training. 
We further investigate these keypoints representations in~\cref{sec:ablation} and present our results in~\cref{tab:ablation}.

\subsection{RotEst}

For each object, from the estimated pixel coordinates of the 32 keypoints (the eight corners of the 3D bounding box and the 24 intermediate bounding box keypoints), the RotEst module predicts the object orientation represented as the 6D continuous representation in SO(3)~\citep{zhou2019continuity}. Formally, the RotEst module takes a 64-dimensional vector (32 pixel coordinates) and generates a 6D object orientation estimate. We implement the RotEst module using six fully connected layers with hidden dimension 1024 and dropout probability of 0.5.
\subsection{Loss Function}
The Hungarian loss consists of four components: class probability loss, bounding box loss, keypoint loss, and pose loss. 
The class probability loss and the bounding box loss follow the DETR model~\citep{carion2020end}.

\subsubsection{Class Probability Loss}
The class probability loss function is the standard negative log-likelihood. Since we choose the cardinality of the set to be higher than the expected maximum number of
objects in an image, the \text{\O} class appears disproportionately often. Thus, we weigh the loss for the \text{\O} class with a factor of 0.4.

\subsubsection{Bounding Box Loss}
We use a weighted combination of the Generalized IoU (GIoU)~\citep{rezatofighi2019generalized} and $\ell_1$-loss with 2 and 5 factors, respectively, for the bounding box loss.

\subsubsection{Keypoint Loss}
\label{sec:kploss}

Having the groundtruth $K_i$ and the model output $\hat{K}_{\hat{\sigma}(i)}$ our keypoints loss can be represented as: 
\begin{equation}\label{eqn:keypoints_loss}
    \mathcal{L}_{keypoints}(K_i, \hat{K}_{\hat{\sigma}(i)}) = \gamma ||K_i - \hat{K}_{\hat{\sigma}(i)}||_1 + \delta \mathcal{L_{CR}},
\end{equation}
where $\gamma$ and $\delta$ are hyperparameters. The first part of the keypoints loss is the $\ell_1$ loss, and for the second part, we employ the cross-ratio loss $\mathcal{L_{CR}}$ provided in Equation~\ref{equ:cr_loss} to enforce the cross-ratio consistency in the keypoints loss as proposed by~\citet{Li_2021_CVPR}. This loss is self-supervised by preserving the cross-ratio of each line to be 4/3. The reason is that after camera projection of the 3D bounding box on the image plane, the cross-ratio of every four collinear points remains the same. 

\begin{equation}
\label{equ:cr_loss}
    \mathcal{L_{CR}} = Smooth\ell_1(\text{CR}^2 -  \frac{||c-a||^2||d-b||^2}{||c-b||^2||d-a||^2}),
\end{equation}
where $\text{CR}^2$ is chosen since $||.||^2$ can be easily computed using vector inner product. Given four collinear points A, B, C and D and their predicted 2D projections a, b, c, and d, the groundtruth cross-ratio CR is defined as:
\begin{equation}
    \text{CR} = \frac{||C-A||~||D-B||}{||C-B||~||D-A||} = \frac{4}{3}.
\end{equation}
\subsubsection{Pose Loss}
We supervise the rotation $R$ and the translation $t$ individually via employing PLoss and SLoss from~\citep{xiang2017posecnn} for rotation and $\ell_1$ loss for translation:

\begin{equation}\label{eqn:ploss}
\mathcal{L}_{pose}(R_i, t_i, \hat{R}_{\sigma(i)}, \hat{t}_{\sigma(i)}) = \mathcal{L}_{rot}(R_i, \hat{R}_{\sigma(i)}) + || t_i - \hat{t}_{\sigma(i)} ||_1,
\end{equation}

\begin{equation}\label{eqn:pose_loss}
\mathcal{L}_{rot} = \left\{\begin{array}{ll}
\frac{1}{|\mathcal{M}_i|} \displaystyle\sum_{\text{x}_1 \in \mathcal{M}_i}  \min_{\text{x}_2 \in \mathcal{M}_i}|| R_i\text{x}_1 - \hat{R}_{\sigma(i)} \text{x}_2 ||_1 \text { if sym, } \\
\frac{1}{|\mathcal{M}_i|} \displaystyle\sum_{\text{x} \in \mathcal{M}_i} || R_i\text{x} - \hat{R}_{\sigma(i)} \text{x} ||_1 \text { otherwise, } 
\end{array}\right.
\end{equation}
where $\mathcal{M}_i$ indicates the set of 3D model points. Here, we subsample 1.5K points from meshes provided with the dataset. $R_i$ is the groundtruth rotation, and $t_i$ is the groundtruth translation. $\hat{R}_{\sigma(i)}$ and $\hat{t}_{\sigma(i)}$ are the predicted rotation and translation, respectively.

\subsection{Model Architecture}
The proposed YOLOPose architecture is inspired by T6D-Direct~\citep{arash2021gcpr}.
The model consists of a ResNet backbone followed by an encoder-decoder based Transformer and MLP prediction heads to predict a set of tuples described in~\cref{sec:set}. CNN architectures have several inductive biases designed into them~\citep{lecun1995convolutional, CohenS17}. These strong biases enable CNNs to learn efficient local spatial features in a fixed neighborhood defined by the receptive field to perform well on many computer vision tasks. 
In contrast, Transformers, aided by self-attention, are suitable for learning spatial features over the entire image. 
This makes the Transformer architecture ideal for multi-object pose estimation. %
In this section, we describe the individual components of the YOLOPose architecture.
\subsubsection{Backbone Network}
We use a ResNet50 backbone for extracting features from the given RGB image. For an image size of height H and width W, the backbone network extracts 2048 low-resolution feature maps of size 
H/32$\times$W/32. We then use 1$\times$1 convolution to reduce the 2048 feature dimensions to a lower number of \textit{d} dimensions. The standard Transformer models are designed to process vectors. Therefore, to enable processing the \textit{d}$\times$H/32$\times$W/32 features, we vectorize them to \textit{d}$\times\frac{H}{32}\frac{W}{32}$.
\subsubsection{Positional Encodings}
Multi-head self-attention, the core component of the Transformer model, is permutation-invariant. Thus, the Transformer architecture ignores the
order of the input vectors. We employ the standard solution of supplementing the input vectors with absolute positional encoding following~\citet{carion2020end} to provide the Transformer model with spatial information of the pixels.
The positional embeddings are added elementwise to the backbone feature vectors before feeding them to the Transformer encoder as input. 

\subsubsection{Encoder}
The Transformer encoder module consists of six encoder layers with skip connections. Each layer performs multi-head self-attention of the input vectors.
The embeddings used in our model are 256-dimensional vectors.
\subsubsection{Decoder}
From the encoder output embedding and $N$ positional embedding inputs, the decoder generates $N$ output embeddings using the multi-head self-attention and cross-attention mechanisms, where $N$ is the cardinality of the predicted set. Unlike the fixed positional encoding used in the encoder, we use learnable positional encoding in the decoder, called \textit{object queries}. From the $N$ decoder output embeddings, we use feed-forward prediction heads
to generate a set of $N$ output tuples independently. 
\subsubsection{FFN}
For each object query decoder output, we use four feed-forward prediction heads shared across object queries to predict the class probability, bounding box, translation, and keypoints independently. Prediction heads are straightforward three-layer MLPs with hidden dimension 256 in each layer and ReLU activation.

\section{Evaluation}
In this section, we evaluate the performance of our proposed YOLOPose model and compare it with other state-of-the-art 6D pose estimation methods.

\begin{figure*}[h]
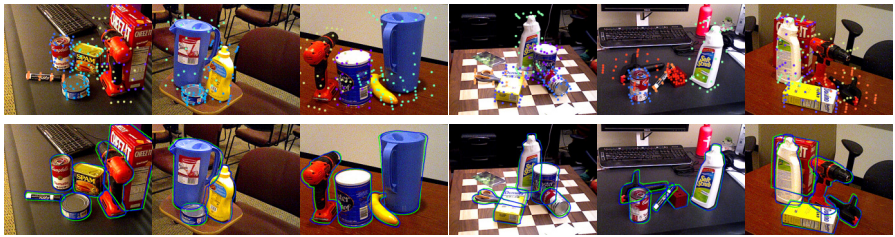

        \centering
        \newlength{\imgres}
        \setlength{\imgres}{0.16\textwidth}
        \setlength{\tabcolsep}{0.01cm}
        \begin{tabular}{cccccc}
            
         \includegraphics[width=\imgres]{figures/results/fig4-13.png} &
         \includegraphics[width=\imgres]{figures/results/fig4-18.png} &
         \includegraphics[width=\imgres]{figures/results/fig4-15.png} &
         \includegraphics[width=\imgres]{figures/results/fig4-16.png} &
         \includegraphics[width=\imgres]{figures/results/fig4-17.png} &
         \includegraphics[width=\imgres]{figures/results/fig4-14.png} \\
            
        \includegraphics[width=\imgres]{figures/results/fig4-4.png} &
         \includegraphics[width=\imgres]{figures/results/fig4-9.png} &
         \includegraphics[width=\imgres]{figures/results/fig4-8.png} &
         \includegraphics[width=\imgres]{figures/results/fig4-10.png} &
         \includegraphics[width=\imgres]{figures/results/fig4-7.png} &
         \includegraphics[width=\imgres]{figures/results/fig4-6.png} \\
        \end{tabular}
        \caption{Qualitative results on YCB-V test set. 
        Top row: The predicted IBB keypoints overlaid on the input images.
        Bottom row: Groundtruth and predicted object poses visualized as object contours in green and blue colors, respectively.}
        \label{fig:result}
\end{figure*}

\subsection{Dataset}

We use the challenging YCB-Video (YCB-V)~\citep{xiang2017posecnn} dataset to evaluate the performance of our model.
YCB-V provides bounding box, segmentation, and 6D pose annotation for 133,936 RGB-D images.
Since our model is RGB-based, we do not use the provided depth information. 
The dataset is generated by capturing video sequences of a random subset of objects from a total of 21 objects placed in tabletop configuration.
There are 92 video sequences, out of which twelve are used for testing and the rest are used for training.
The objects used exhibit varying geometric shapes, reflectance properties, and symmetry. Thus, YCB-V is a challenging dataset for benchmarking
6D object pose estimation methods.
YCB-V also provides high-quality meshes for all 21 objects.
Mesh points from these objects are used in computing the evaluation metrics discussed in~\cref{sec:metric}.~\citet{hodavn2020bop} provided a variant of YCB-V\footnote{\url{https://bop.felk.cvut.cz/datasets/}} as a part of the BOP challenge in which the centers of the 3D bounding boxes are aligned with the origin of the model coordinate system, and the groundtruth annotations are converted correspondingly. We use the BOP variant of the YCB-V dataset.
Additionally, we use the COCO dataset~\citep{lin2014microsoft} for pretraining our model on the task of object detection.

\subsection{Metrics}
\label{sec:metric}

\citet{xiang2017posecnn} proposed area under the curve (AUC) of the ADD and ADD-S metrics for evaluating the accuracy of non-symmetric and symmetric objects, respectively.
Given the groundtruth 6D pose annotation with rotation and translation components $R$ and $t$, and the predicted rotation and translation components 
$\hat{R}$ and $\hat{t}$, the ADD metric is the average $\ell_2$ distance between the subsampled mesh points $\mathcal{M}$ in the groundtruth and the predicted pose. 
In contrast, the symmetry aware ADD-S metric is the average distance between the closest subsampled mesh points $\mathcal{M}$ in the groundtruth and predicted pose.
A predicted pose is considered correct if the ADD metric is below 0.1\,m. The ADD(-S) metric is the combination of ADD-S for symmetric objects and ADD for the non-symmetric objects. Formally, 

\begin{equation}
        \text{ADD} = \frac{1}{|\mathcal{M}|} \sum_{x \in \mathcal{M}}\|(Rx+t)-(\hat{R} x+\hat{t})\|,
\end{equation}

\begin{equation}
        \text{ADD-S} = \frac{1}{|\mathcal{M}|} \sum_{x_{1} \in \mathcal{M}} \min _{x_{2} \in \mathcal{M}}\|(R x_{1}+t)-(\hat{R} x_{2}+\hat{t})\|.
\end{equation}

\subsection{Hyperparameters}
The $\gamma$ and $\delta$ hyperparameters in $\mathcal{L}_{keypoints}$ (\cref{eqn:keypoints_loss}) are set to 10 and 1, respectively.
While computing the Hungarian loss, the pose loss component is weighted down by a factor 0.02. The cardinality of the predicted set $N$ is set to 20.
The model takes images of the size 640 $\times$ 480 as input and 
is trained using the AdamW optimizer~\citep{adamw}
with an initial learning rate of $2e^{-4}$ for 335K iterations. 
The learning rate is decayed to $2e^{-5}$ after 271K iterations, and the batch size is 32. Moreover, gradient clipping with a maximal gradient norm of 0.1 is applied.
In addition to YCB-V dataset images, we use the synthetic dataset provided by PoseCNN for training our model.

\subsection{Results}
\begin{table}[h]
\centering
\label{tab:ycbv-details}
\setlength{\aboverulesep}{0pt}
\setlength{\belowrulesep}{0pt}
\caption{Comparison of keypoints-based method with state-of-the-art methods on YCB-V. P.E=1 denotes one model for all objects, whereas P.E.=($N$) denotes the usage of object specific models. The symmetric objects are denoted by *, and the best results are shown in bold.}
\label{tab:ycbv-details-keypoints}
\resizebox{\textwidth}{!}{
\begin{tabular}{l|cc|c|cc|cc|cc|cc}
\toprule
Method & \multicolumn{2}{c|}{PoseCNN~\citep{xiang2017posecnn}} &
\multicolumn{1}{c|}{PVNet~\citep{peng2019pvnet}} &
\multicolumn{2}{c|}{GDR-Net~\citep{Wang_2021_GDRN}} &  
\multicolumn{2}{c|}{T6D-Direct~\citep{arash2021gcpr}} &
\multicolumn{2}{c|}{YOLOPose (Ours)} &
\multicolumn{2}{c}{DeepIM~\citep{li2018deepim}} \\
\midrule
P.E. & \multicolumn{2}{c|}{1} &
\multicolumn{1}{c|}{$N$} &
\multicolumn{2}{c|}{1} &  
\multicolumn{2}{c|}{1} &
\multicolumn{2}{c|}{1} &
\multicolumn{2}{c}{1} \\
\midrule
Metric & 
\thead{AUC of \\ ADD-S} & \thead{AUC of \\ ADD(-S)} &
\thead{AUC of \\ ADD(-S)} & \thead{AUC of \\ ADD-S} & 
\thead{AUC of \\ ADD(-S)} & \thead{AUC of \\ ADD-S} & 
\thead{AUC of \\ ADD(-S)} & \thead{AUC of \\ ADD-S} & 
\thead{AUC of \\ ADD(-S)} & 
\thead{AUC of \\ ADD-S} & \thead{AUC of \\ ADD(-S)}\\
\midrule
master\_chef\_can       & 84.0 & 50.9 & 81.6 & 96.6 & 71.1 & 91.9 & 61.5 & 91.3 & 64.0 & 93.1 & 71.2 \\
cracker\_box            & 76.9 & 51.7 & 80.5 & 84.9 & 63.5 & 86.6 & 76.3 & 86.8 & 77.9 & 91.0 & 83.6 \\
sugar\_box              & 84.3 & 68.6 & 84.9 & 98.3 & 93.2 & 90.3 & 81.8 & 92.6 & 87.3 & 96.2 & 94.1 \\
tomato\_soup\_can       & 80.9 & 66.0 & 78.2 & 96.1 & 88.9 & 88.9 & 72.0 & 90.5 & 77.8 & 92.4 & 86.1 \\
mustard\_bottle         & 90.2 & 79.9 & 88.3 & 99.5 & 93.8 & 94.7 & 85.7 & 93.6 & 87.9 & 95.1 & 91.5 \\
tuna\_fish\_can         & 87.9 & 70.4 & 62.2 & 95.1 & 85.1 & 92.2 & 59.0 & 94.3 & 74.4 & 96.1 & 87.7 \\
pudding\_box            & 79.0 & 62.9 & 85.2 & 94.8 & 86.5 & 85.1 & 72.7 & 92.3 & 87.9 & 90.7 & 82.7 \\
gelatin\_box            & 87.1 & 75.2 & 88.7 & 95.3 & 88.5 & 86.9 & 74.4 & 90.1 & 83.4 & 94.3 & 91.9 \\
potted\_meat\_can       & 78.5 & 59.6 & 65.1 & 82.9 & 72.9 & 83.5 & 67.8 & 85.8 & 76.7 & 86.4 & 76.2 \\
banana                  & 85.9 & 72.3 & 51.8 & 96.0 & 85.2 & 93.8 & 87.4 & 95.0 & 88.2 & 91.3 & 81.2 \\
pitcher\_base           & 76.8 & 52.5 & 91.2 & 98.8 & 94.3 & 92.3 & 84.5 & 93.6 & 88.5 & 94.6 & 90.1 \\
bleach\_cleanser        & 71.9 & 50.5 & 74.8 & 94.4 & 80.5 & 83.0 & 65.0 & 85.3 & 73.0 & 90.3 & 81.2 \\
bowl$^*$                & 69.7 & 69.7 & 89.0 & 84.0 & 84.0 & 91.6 & 91.6 & 92.3 & 92.3 & 81.4 & 81.4 \\
mug                     & 78.0 & 57.7 & 81.5 & 96.9 & 87.6 & 89.8 & 72.1 & 84.9 & 69.6 & 91.3 & 81.4 \\
power\_drill            & 72.8 & 55.1 & 83.4 & 91.9 & 78.7 & 88.8 & 77.7 & 92.6 & 86.1 & 92.3 & 85.5 \\
wood\_block$^*$         & 65.8 & 65.8 & 71.5 & 77.3 & 77.3 & 90.7 & 90.7 & 84.3 & 84.3 & 81.9 & 81.9 \\
scissors                & 56.2 & 35.8 & 54.8 & 68.4 & 43.7 & 83.0 & 59.7 & 93.3 & 87.0 & 75.4 & 60.9 \\
large\_marker           & 71.4 & 58.0 & 35.8 & 87.4 & 76.2 & 74.9 & 63.9 & 84.9 & 76.6 & 86.2 & 75.6 \\
large\_clamp$^*$        & 49.9 & 49.9 & 66.3 & 69.3 & 69.3 & 78.3 & 78.3 & 92.0 & 92.0 & 74.3 & 74.3 \\
extra\_large\_clamp$^*$ & 47.0 & 47.0 & 53.9 & 73.6 & 73.6 & 54.7 & 54.7 & 88.9 & 88.9 & 73.3 & 73.3 \\
foam\_brick$^*$         & 87.8 & 87.8 & 80.6 & 90.4 & 90.4 & 89.9 & 89.9 & 90.7 & 90.7 & 81.9 & 81.9 \\
\midrule
MEAN                    & 75.9 & 61.3 & 73.4 & 89.1 & 80.2 & 86.2 & 74.6 & \textbf{90.1} & \textbf{82.6} & 88.1 & 81.9 \\
\bottomrule
\end{tabular}
}
\end{table}

\begin{table}[h]
      \centering
      \footnotesize
      \setlength{\aboverulesep}{0pt}
      \setlength{\belowrulesep}{0pt}
        \caption{Results on YCB-V. 
        }
\begin{tabular}{l|c|c|c|c}
    \toprule
    Method & ADD(-S) & \thead{AUC of \\ ADD-S} & \thead{AUC of \\ ADD(-S)} & \thead{Inference \\Time \\{} [\si{\s}] }\\
    \bottomrule
    CosyPose$^{\dagger}$~\citep{labbe2020} & - & 89.8 & \textbf{84.5} & 0.395\\
    PoseCNN~\citep{xiang2017posecnn} & 21.3 & 75.9 & 61.3 & -\\
    SegDriven~\citep{hu2019segmentation} & 39.0 & - & - & -\\
    Single-Stage~\citep{hu2020single} & 53.9 & - & - & -\\
    GDR-Net~\citep{Wang_2021_GDRN} & 49.1 & 89.1 & 80.2 & 0.065\\
    T6D-Direct~\citep{arash2021gcpr}  & 48.7 & 86.2 & 74.6 & \textbf{0.017}\\
    YOLOPose (Ours) &  \textbf{65.0} & \textbf{90.1} & 82.6 & \textbf{0.017} \\
    \bottomrule
  \end{tabular}
    \label{tab:inftime}
    \\
    $^{\dagger}$ Refinement-based method.
\end{table}

\begin{figure}[h]
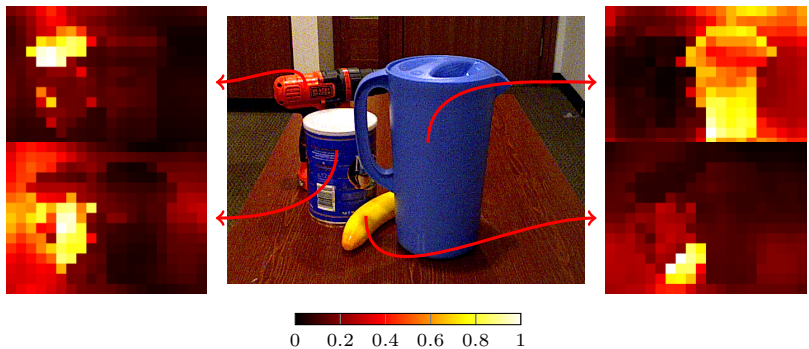

        \centering
          \resizebox{.9\linewidth}{!}{%
\begin{tikzpicture}

\node (scene) at (0, 0) {\includegraphics[width=0.4\linewidth]{viz/fig6-scene.jpg}};

\node (sa1) at (-4.1, 0.93) {\includegraphics[width=0.225\linewidth]{viz/fig6-120-160.jpg}};
\node (sa2) at (-4.1, -0.93) {\includegraphics[width=0.225\linewidth]{viz/fig6-259_172.jpg}};
\node (sa3) at (4.1, 0.93) {\includegraphics[width=0.225\linewidth]{viz/fig6-192-384.jpg}};
\node (sa4) at (4.1, -0.93) {\includegraphics[width=0.225\linewidth]{viz/fig6-364-249.jpg}};

\coordinate (sa1_pixel) at (-1.375, 0.85);
\coordinate (sa2_pixel) at (-0.95, .0);
\coordinate (sa3_pixel) at (0.3, 0.1);
\coordinate (sa4_pixel) at (-0.55, -0.9);

\draw [very thick, ->, red]    (sa1_pixel) to[out=90, in=0] (sa1.east);
\draw [very thick, ->, red]    (sa2_pixel) to[out=-90, in=0] (sa2.east);
\draw [very thick, ->, red]    (sa3_pixel) to[out=90, in=180] (sa3.west);
\draw [very thick, ->, red]    (sa4_pixel) to[out=-90, in=180] (sa4.west);

\end{tikzpicture}
}
          \tikz{
                \pgfplotscolorbardrawstandalone[ 
                    colormap/hot2,
                    point meta min=0,
                    point meta max=1,
                    colorbar horizontal, 
                    colorbar style={
                        height=0.15cm,
                        width=3cm,
                        font=\scriptsize
                }]
            }
        \vspace{-2mm}
          \caption{Encoder self-attention. We visualize the self-attention maps for four pixels belonging to four objects
          in the image. Note that for each object, roughly all the corresponding pixels are attended.}
          \label{fig:enc_sa}
\end{figure}

\begin{figure}[h]
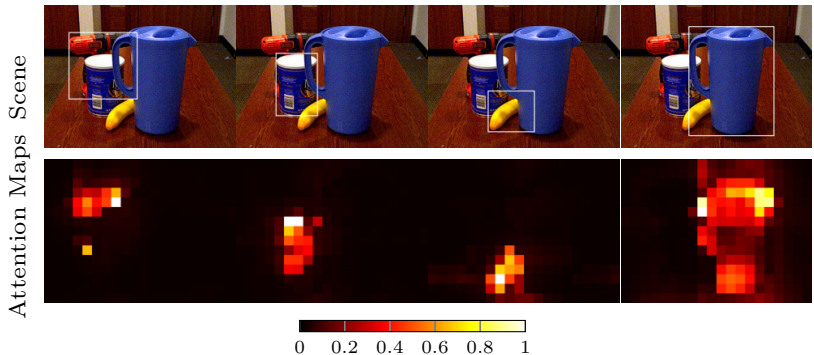

        \centering
        \newlength{\imgw}
        \setlength{\imgw}{2.0cm}
        \setlength{\tabcolsep}{0.009cm}
        \resizebox{0.9\linewidth}{!}{
        \begin{tabular}{p{1.em}cccc}
                \raisebox{2.75\normalbaselineskip}[0pt][0pt] {\rotatebox[origin=c]{90}{\scriptsize Scene \qquad \qquad}} &
         \includegraphics[width=\imgw]{viz/fig5-1-1.jpg} &
         \includegraphics[width=\imgw]{viz/fig5-1-2.jpg} &
         \includegraphics[width=\imgw]{viz/fig5-1-3.jpg} &
         \includegraphics[width=\imgw]{viz/fig5-1-4.jpg} \\
         \raisebox{1.75\normalbaselineskip}[0pt][0pt] {\rotatebox[origin=c]{90}{\scriptsize Attention Maps}} &
         \includegraphics[width=\imgw]{viz/fig5-2-1.jpg} &
         \includegraphics[width=\imgw]{viz/fig5-2-2.jpg} &
         \includegraphics[width=\imgw]{viz/fig5-2-3.jpg} &
         \includegraphics[width=\imgw]{viz/fig5-2-4.jpg} \\
        \end{tabular}
        }
        \tikz{
            \pgfplotscolorbardrawstandalone[ 
                colormap/hot2,
                point meta min=0,
                point meta max=1,
                colorbar horizontal, 
            colorbar style={
                height=0.15cm,
                width=3cm,
                font=\scriptsize
            }]
        }
        \caption{Top: Object detections predicted by bounding boxes in a given image.
        Bottom: Decoder cross-attention maps for the object queries corresponding to the predictions in the first row.
        }
        \label{fig:multiheadattn}
\end{figure}
In this section, we present the quantitative and qualitative results of our method.
In~\cref{tab:ycbv-details}, we provide the quantitative per class area under the accuracy curve (AUC) of the ADD-S and ADD(-S) metrics. 
Except for DeepIM, a refinement-based method and PVNet, an indirect method, all other methods estimate the 6D pose directly.
Our method outperforms all of the competing approaches.
Additionally in~\cref{tab:inftime}, we present Average Recall (AR) of ADD(-S) 0.1d, and AUC of ADD-S and ADD(-S) of the state-of-the-art methods. In terms of the AR of ADD(-S) and AUC of ADD-S metrics, our method achieves state-of-the-art results among the pose estimators.
Note that the pose refinement approach, CosyPose, achieves the best result in terms of the AUC of ADD(-S) metric. However, in terms of the approach pose refinement methods are orthogonal to the pose estimation methods and can benefit from the improved pose estimation accuracy. Furthermore, pose estimation models with admissible accuracy avoid the need for training an additional pose refinement model and enable faster inference time.
We present exemplar qualitative results in~\cref{fig:result}. In~\cref{fig:enc_sa}, we visualize encoder self-attention of four different pixels
belonging to four different objects and in~\cref{fig:multiheadattn}, we visualize the decoder cross-attention corresponding to four different object detections.
In both visualizations, the attended regions correspond to the spatial position of the object in the image very well.

\subsection{Inference Time Analysis}
In terms of the inference speed, one of the major advantages of our architecture is that the feed-forward prediction networks (FFN) can be executed in parallel
for each object. Thus, irrespective of the number of objects in an image, our model generates pose predictions in parallel. In~\cref{tab:inftime}, we present
the inference time results for 6D pose estimation. Our method operates at $\sim$59 FPS on an NVIDIA 3090 GPU and Intel 3.70\,GHz CPU and is, hence, suitable for real-time applications.

\section{Ablation Study}\label{sec:ablation}
In contrast to the standard approach of estimating the 2D keypoints and using P\textit{n}P solver---which is not trivially differentiable---to estimate the 6D object pose, 
we use the learnable RotEst module to estimate the object orientation from a set of predicted interpolated keypoints.
In this section, we analyze the effectiveness of our RotEst module and the choice of the keypoint representation.
\subsection{Effectiveness of Keypoints Representations}

We compare various keypoint representations, namely 3D bounding box keypoints (BB), random keypoints sampled using the FPS algorithm and our representation of choice, the interpolated bounding box keypoints (IBB).
We use the OpenCV implementation of the RANSAC-based EP\textit{n}P algorithm with the same parameters to recover 6D object pose from the predicted keypoints. Since EP\textit{n}P does not contain any learnable components, this experiment serves the goal of evaluating the ability of the YOLOPose model to estimate different keypoint representations in isolation.
YOLOPose is trained using only the $\ell_1$ loss in the case of BB and FPS representations, 
whereas in the case of IBB representation, $\ell_1$ is combined with the cross-ratio loss described in~\cref{sec:kploss}. 
In our experiments presented in~\cref{tab:ablation}, when used in conjunction with the EP\textit{n}P solver, the FPS keypoints performed worse than all other representations. 
In contrast, the IBB keypoints representation yields the best performance. We conjecture that the additional cross-ratio loss employed helps our model in learning the IBB keypoint projections better.

\subsection{Effectiveness of RotEst}
After deciding on the keypoint representation, we compare the performance of the learnable feed-forward rotation and translation estimators against 
the analytical EP\textit{n}P algorithm.
Based on the observation that the rotation and translation components impacted by different factors~\citep{li2019cdpn}, 
we decide to estimate rotation and translation separately. 
As shown in~\cref{tab:ablation}, using only the rotation from the EP\textit{n}P result and directly regressing the translation improved the accuracy significantly.
In general, RotEst performs slightly better than using EP\textit{n}P orientation and direct translation estimation. 
Furthermore, the RotEst module and the translation estimators are straightforward MLPs and, thus, do not add much execution time overhead. 
This enables YOLOPose to perform inference in real-time.

\begin{table}[h]
  \centering
  \footnotesize
  \setlength{\aboverulesep}{0pt}
  \setlength{\belowrulesep}{0pt}
  \caption{Ablation study on YCB-V. We present the comparison results of different keypoint representations and the effectiveness of RotEst. 
    The top section of the table corresponds to different keypoint representations in combination with the standard EP\textit{n}P algorithm, and 
    the bottom section corresponds to the effectiveness of the learnable RotEst module using IBB keypoints.}
  \begin{tabular}{l|c|c}
    \toprule
     Method & ADD(-S) & \thead{AUC of \\ ADD(-S)} \\
    \midrule
    FPS + EP\textit{n}P & 31.4 & 56.9 \\
    handpicked + EP\textit{n}P  & 31.5 & 55.7 \\
    IBB + EP\textit{n}P & \textbf{56.0} & \textbf{74.7} \\
    \midrule
    IBB + EP\textit{n}P for $R$; head for $t$ & 63.9 & 82.3 \\
    IBB + RotEst for $R$ and head for $t$ & \textbf{65.0} & \textbf{82.6} \\
    \bottomrule
  \end{tabular}
  \label{tab:ablation}
\end{table}

\section{Discussion \& Conclusion}
We presented YOLOPose, a Transformer-based single-stage multi-object pose estimation method using keypoint
regression. Our model jointly estimates bounding boxes, class labels, translation vectors, and pixel coordinates of 
3D keypoints for all objects in the given input image.
Employing the learnable RotEst module to estimate the object orientation from the predicted keypoints coordinate enables the model to be end-to-end differentiable. We evaluated our model on the widely-used YCB-Video dataset and reported results comparable to the state-of-the-art approaches while being real-time capable. In the future, we plan to extend our model to video sequences and exploit temporal consistency to improve the pose estimation accuracy further.

\section{Acknowledgment}
This work has been funded by the German Ministry of Education and Research (BMBF), grant no. 01IS21080,
project ``Learn2Grasp: Learning Human-like Interactive Grasping based on Visual and Haptic Feedback''.

\bibliography{references}

\end{document}